# Utilization of Multinomial Naive Bayes Algorithm and Term Frequency-Inverse Document Frequency *(TF-IDF Vectorizer)* in Checking the Credibility of News Tweet in the Philippines

**Author: Neil Christian R. Riego**
College of Engineering and Technology
Pamantasan ng Lungsod ng Maynila
*General Luna, corner Muralla St,*
*Intramuros, Manila, 1002 Metro Manila*
neilchristianriego3@gmail.com

**Author: Danny Bell Villarba**
College of Engineering and Technology
Pamantasan ng Lungsod ng Maynila
*General Luna, corner Muralla St,*
*Intramuros, Manila, 1002 Metro Manila*
dannybellvillarba187@gmail.com

***Abstract*** *- The digitalization of news media become a good indicator of progress and signal to more threats. Media disinformation or fake news is one of these threats, and it is necessary to take any action in fighting disinformation. This paper utilizes ground truth-based annotations and TF-IDF as feature extraction for the news articles which is then used as a training data set for Multinomial Naive Bayes. The model has an accuracy of 99.46% in training and 88.98% in predicting unseen data. Tagging fake news as real news is a concerning point on the prediction that is indicated in the F1 score of 89.68%. This could lead to a negative impact. To prevent this to happen it is suggested to further improve the corpus collection, and use an ensemble machine learning to reinforce the prediction*

***Index Terms*** *- TF-IDF, Philippines' Election, News Credibility, Fake News, Multinomial Naive Bayes, Twitter, Machine Learning*

## 1. INTRODUCTION

### *1.1. Research Story*

Media is considered an influential entity across multiple fields of concern. Though media alone is impactful enough to society, additional influence from highly regarded parties can cause indirect mass manipulation. In Facebook, the effect is most substantial when politicians act as intermediary-senders. It is influenced by the party affiliation of the intermediary-sender and the social media audience [1].

On the other hand, social media influencers are groups or individuals whose public appeal is higher than ordinary people. A 2019 study reported in Adweek found that 54.7% of respondents follow influencers to learn new things, 53.1% want to learn more about what they do, and 51.6% to get inspiration for their own life [2]. Despite the fragile reputation a person can exhibit within the digital community, some social media influencers act in a play-safe manner. On the contrary, such action observed can manipulate or influence others.

In addition, Twitter is a microblogging type of social media that emphasizes threads of information and provides a certain degree of freedom to its users in terms of their subjective expressions The platform contributes to the risk of its users being exposed to wrong interpretation and understanding of information.

In a study by Lee and Lindsey in 2017, Participants were more likely to expose themselves to specific news sources selectively, possible sources they trust more than others when they felt highly overloaded with news information received on social media [3]. This may imply the role of social media influencers in Twitter, which can be deemed as a double-edged sword. They can help fight misinformation or make it worse by mixing their influence and opinion in a sensitive argument.

### *1.2. Problem Statement*

Fake news can negatively impact society due to the growing use of mobile devices and the global increase of internet access [4]. The pandemic caused state leaders to resort to social media platforms that may induce a negative impact on the local state integrity due to the spread of fake news.

The diversity of social media users ranging from minority groups or individuals to highly regarded personalities and influencers are all susceptible to the risk of possible information fabrication or fake news caused by uneducated users. Even iconic individuals may cause misinformation by being more appealing when information overload occurs to the masses, and the people favor influenced insights more than profound sources [3].

### *1.3. Related Works*

Despite that fake news is one of the contemporary problems that are encountered in a global scale manner, it was an unforeseen case that only a few literary studies addressing the formulation of automated solutions were published.

The data acquisition process also utilizes a weak unsupervised data processing algorithm as discussed by Helmstetter and Paulheim in 2021, the TD-IDF is used in tweet classification of large-scale data sets retrieved from Twitter API and external crowdsourced organizations [6]. It also utilizes ground truth based on reliable sites and datasets, which is adopted also in this study. Additionally, the vector-matrix can be applied into a support vector machine or Multinomial Naive Bayes as seen in the study of Alonso, Vilares, Gómez-Rodríguez, and Vilares are resorting to a supervised data processing approach integrated with a ranking support vector machine algorithm for the analysis of tweets credibility [7].

The study of Cruz, Tan, and Cheng in 2020 inhibits complex Multi Transfer Learning (MTL) techniques to enhance fake news classification accuracy. The data processing algorithm was applied to Filipino news articles that are published in their native Tagalog language. [8] Furthermore, the study of Villavicencio, Macrohon, Inbaraj, Jeng, and Hsieh in 2021 introduced the concept incorporating NLP techniques, sentiment analysis, and Naive Bayes addressing the covid 19 vaccines in the Philippines [9].

Alonso, Vilares, Gómez-Rodríguez, and Vilares in 2021 applied the sentiment analysis to serve as a complementary process for boosting the effectiveness and accuracy of results from automated fake news detection systems [10]. Similar in how SVM can also be used to supplement the Multinomial Naive Bayes Model.

As such, the related articles above contribute to the similar goal that the study aims to achieve, which is to utilize the functionality of multinomial Naive Bayes algorithm and TF-IDF vectorizer to achieve credibility checking in a different algorithmic concept.

## 2. METHODOLOGY

The study shall engage credibility checking of the news whilst based on the data mining process sequence as depicted in the following Figure 1.

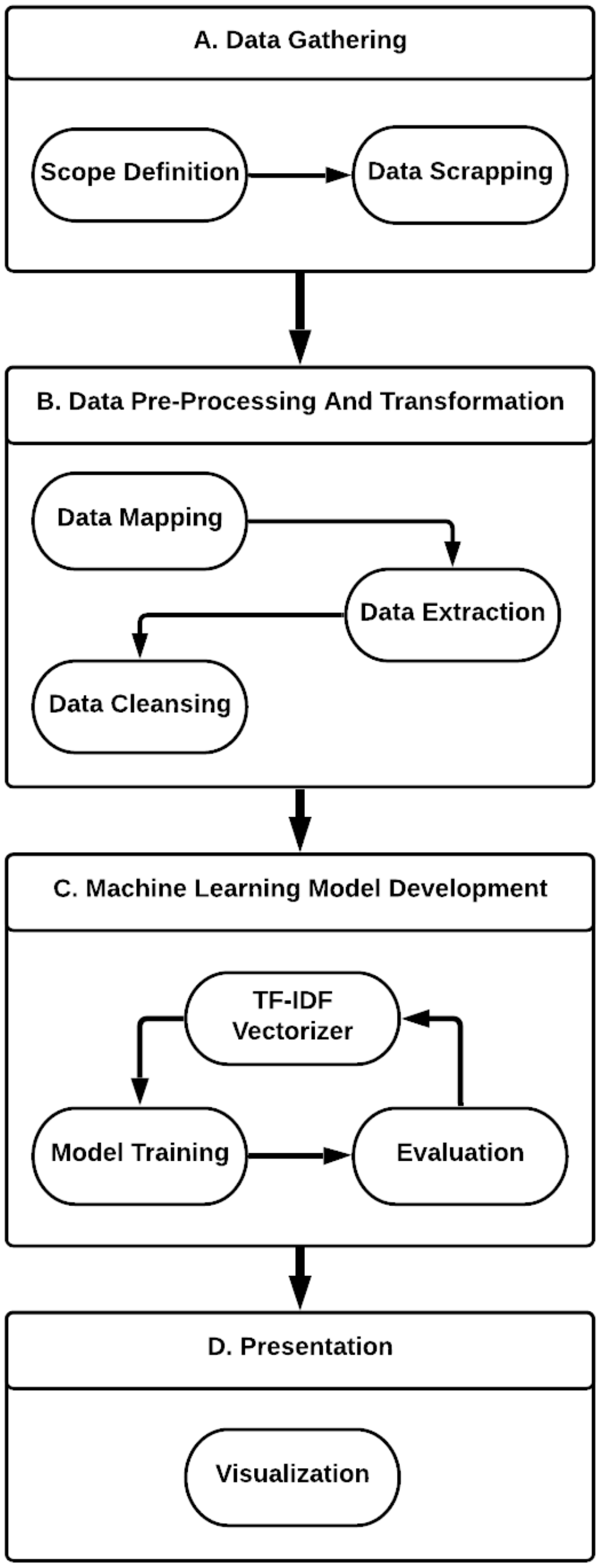

**Figure 1. Proposed Data Mining Process Structure**

The figure above depicts the data mining process structure observed by the researchers throughout the whole study.

### *2.1. Data Gathering*

The dataset is acquired through data scraping on the Twitter platform using the Rapid Miner software. Labels of the scraped data were based on the ground truth based on the list of reliable sources found online.

#### A. Scope Definition

To ensure the quality of data gathered from Twitter the following scope and qualifications are defined as:

1. Data collected from Twitter must be from users with the following qualifications:
    a. Users must be Filipino regardless of their current location of residence;
    b. Users associated with the field of news and politics, and/or;
    c. Accounts that are used for fact-checking and news sharing.
2. Tweets collected must not be:
    a. A retweet coming from other users;
    b. Tweeted before October 1, 2021 [11], *with an exception to tsekph that fact-checks only Philippines' Elections related news* [12].
    c. Any form of a tweet; must be news and tweeted during the start of COC filing up to the Philippines' Election.
3. Scope of news tweet/dataset must be from credible sources [6] such as:
    a. Journalist or reporters;
    b. Breaking or general news brand [13];
    c. Digital publications, and;
    d. Corpus validated under news topic [8].

### B. Data Scraping

Data scraping was conducted through collections of credible sources. Tweets from credible sources such as verified news and government channels are labeled as real news [6]. The following Twitter accounts are the top five (5) trustworthy based on survey results from the 2021 Digital News Report [14]:

1. GMA News Online (@gmanews)
2. ABS-CBN News Online (@ANCALERTS)
3. Philippine Daily Inquirer Online (@phildailyinq)
4. Rappler (@rapplerdotcom)
5. Yahoo! News (@YahooNews)

Fact-checking accounts scan the internet for news and label it as factual, fake, misleading, or it needs additional context. The study considers this as a ground truth since it is validated by a user and/or service. The following Twitter accounts are the two accounts used in the study:

1. tsekph (@tsekph) [12]
2. VERA Files (@verafiles)

## *2.2. Data Pre-processing and Transformation*

The team conducted the following sequence of steps to gather data from multiple sources, process its raw information, and transform it into a useful and efficient format dataset.

### A. Data Mapping

Data mapping was done by semi-manually categorizing the gathered data as fake or real news to arrange and connect each different source into several datasets and later on will be merged on a centralized dataset that contains only the characteristics that are relevant to all sources.

### B. Data Extraction

Datasets were gathered through a combination of data mining software called RapidMiner, Twitter API, and Online Sources. Most news articles were gathered from RapidMiner while Twitter resources were mainly retrieved from Twitter API. Tagalog news corpus that was produced by Cruz, Tan, and Cheng in 2020 using Multi Transfer Learning (MTL) was also used [8].

- ***Dataset after Data Mapping and Extraction***

    The number of observations of the dataset after data extraction reaches 1161 observations with a distribution of the following: news (500 tweets), verafiles (415 tweets), tsekph (246 tweets), with the features of label and article. In addition to this, there are 3206 observations from the Tagalog News Corpus [8].

### C. Data Cleansing

Data were cleaned through the built-in data cleaning features of RapidMiner software to eliminate redundant and irrelevant elements of the datasets.

In cleansing the data, we used several operators in RapidMiner such as filter, transform, and replace operators. The filter operator is mainly used to filter out the rows with retweets, and common words, while the transform operator is used to lowercase all of the data gathered. The replace operator is used to remove some strings such as hashtags, mentions, and links. The following regular expression is used:

**HTTPS Links:** https?://[-a-zA-Z0-9+&@#/%?=~_|!:,.;]*[-a-zA-Z0-9+&@#/%=~_|]\s?

**Twitter Links:** https?://(t.co/)?[-a-zA-Z0-9/]*\s?

**Headlines (tsekph):** #?[fF]act[-\s]?[cC]heck\sby\s?:?\s?(([-a-zA-Z0-9-  ]*)?(@[a-zA-Z0-9]{4,15})?)\s?:?\s?((FALSE)?([fF]alse)?(MISLEADING)?([Mm]isleading)?(NEEDS CONTEXT)?([Nn]eeds context)?([Nn]o basis)?(NO BASIS)?)\s?

**Hashtag:** #[-!"#$%&'()*+,./:;<=>?@\[\\\]_`{|}~\w]*

**Mention:** @[-a-zA-Z0-9_]{4,15}

Furthermore, **the** text pre-processing is done through **RStudio** and RapidMiner. Tokenization, segmentation,

stemming, and filtering of stop words. We also used Tagalog stop words to further reduce the features and tokens based on Tagalog words. This is contributed by Gene Diaz which is uploaded through GitHub [15].

- ***Dataset after Data Cleansing***

The number of observations of the dataset after data cleansing is trimmed down to 744 observations with a distribution of the following: news (431 tweets), verafiles (97 tweets), tsekph (216 tweets). In addition to this, the 3206 observations from the Tagalog News Corpus are trimmed down to the distribution of 1509 real news and 1496 fake news.

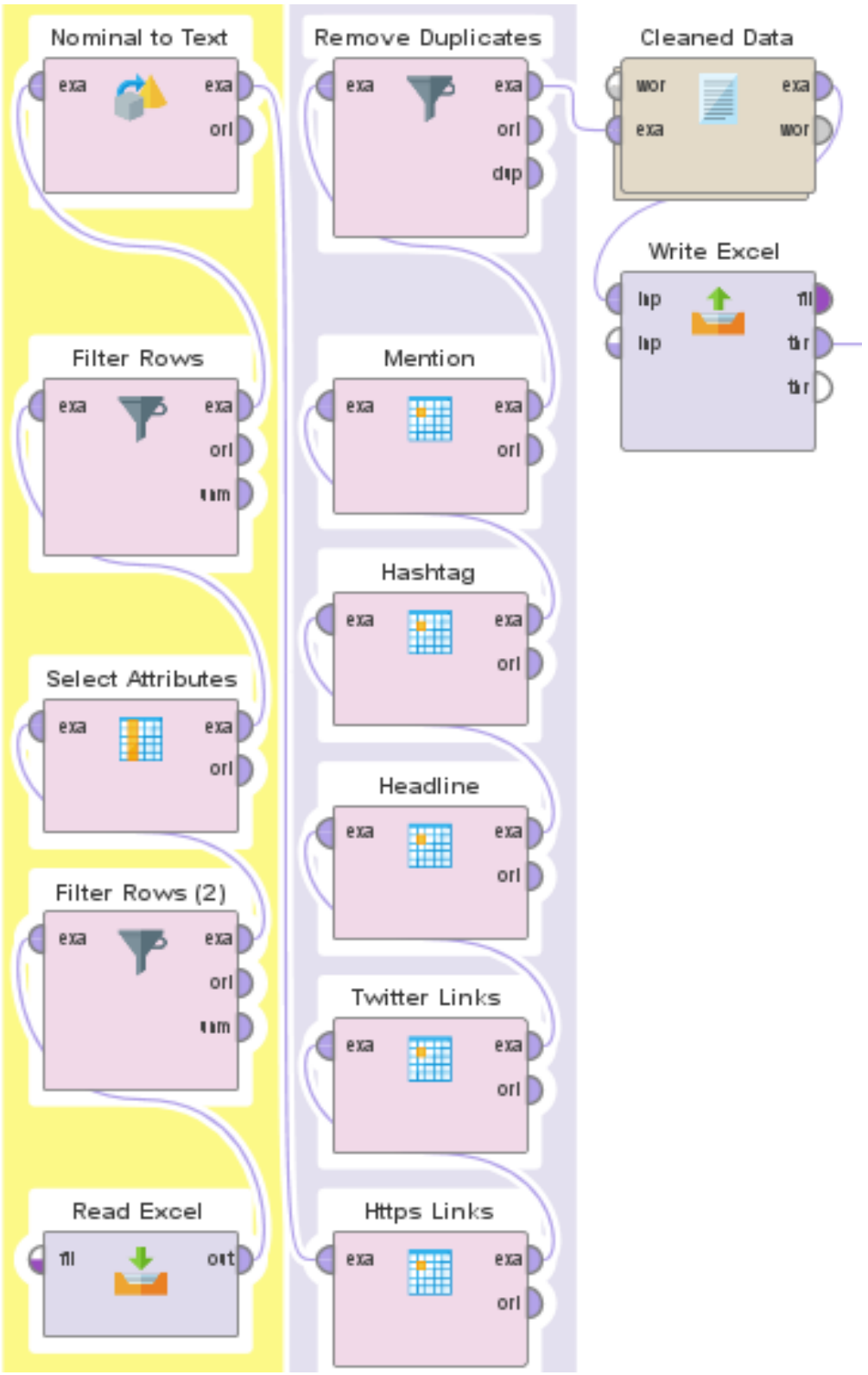

**Figure 2. Data Pre-processing through RapidMiner**

Figure 2 shows the process of data pre-processing through RapidMiner. This is done using the methods stated above.

### *2.3 Machine Learning Model Development*

The model development for machine learning was based on a cycle of what data was identified and what features were to be extracted and then use to develop the model. Afterward, the model shall be evaluated and may restart the development cycle again if there are modifications.

- ***Dataset before Feature Extraction***

The number of observations of the dataset before feature extraction is trimmed down to 790 with an equal distribution of 395 fake and real news. 100 observations per class is coming from the Tagalog corpus to reinforce the corpus with Tagalog articles. For the fake news, the 313 observations are trimmed down to 295 after the cleansing on the merged dataset. To set an equal distribution, we randomly select 295 observations from the real news dataset.

**Table 1: Sample Dataset before Feature Extraction**

| Label | Article |
|---|---|
| real | said difficult impose wealth tax instead suggested progressive tax system undo effects train law live |
| real | lacson areas baka mag reclaim naka base siyensiya datos ba talaga ba makakasama environment ibang lugar naman puwede |
| fake | isinalaysay sumukong suspek pagpatay christine lee silawan krimen ayon renato llenes pekeng facebook account maka chat silawan magka sintahan lang chat march magkita silawan lapu lapu city … |
| fake | social security act nilikha senador richard gordon nilagdaan pangulong duterte nuong pebrero bukos layunin tulungan miyembro oras paghahanap bagong mapapasukan layon nitong palakasin regulatory powers bsp silang makakuha buwanang … |
| fake | icymi pink light display month commemorates support president suggested fb page photo tokyo landmark read |

Table 1 is the preview of the dataset used in the study after the preprocessing which will be further tokenized using feature extraction, specifically TF-IDF vectorizer.

## A. Term Frequency - Inverse Document Frequency

Term Frequency - Inverse Document Frequency (TF-IDF) is used to find the weight of each term in the document. The method focuses on evaluating how the relevance of words across sets of data. TF-IDF is calculated using the formula below.

$$\textbf{TF IDF (t, d) = TF(t, d) * IDF(t)}$$

$$\textbf{TF (t, d)} = \left(\frac{freq(t,d)}{\sum_i^n \ freq(t_i,d)}\right)$$

$$\textbf{IDF(t) = log} \left(\frac{N}{count(t)}\right)$$

**Where:**
$freq(t,d)$ be the count of the instances of the term t in document d.
$count(t)$ be the count of documents in the corpus in which the term t is present

## B. Model Training

The model used for training was the Multinomial Naive Bayes (MNB). MNB is one of the significant learning classifications whose main function is to be used for the analysis of categorical text data. MNB is similar to Simple Naive Bayes which focuses on the independence of each of the model features but uses multinomial distribution for each feature instead. MNB uses the following formula for calculation:

$$P(A|B) = P(A)*P(B|A)/P(B)$$

***Where*** *we are calculating the probability of class A when predictor B is already provided.*

**P(B) =** prior probability of B
**P(A) =** prior probability of class A
**P(B|A) =** occurrence of predictor B given class A probability

## C. Evaluation

Evaluation of the developed model was based on a confusion matrix which is governed by the following formulas:

**Accuracy (all correct / all)** = TP + TN / TP + TN + FP + FN
**Precision (true positives / predicted positives)** = TP / TP + FP
**Sensitivity aka Recall (true positives / all actual positives)** = TP / TP + FN
**F1 Score**= 2TP / (2TP + FP + FN)

### *2.4 Data Visualization*

During the data pre-processing, some data exploration is implemented. This is also relevant when it comes to presenting the developed ML model this is to further show visually the results of the training. Geometric histogram and cloud of words are some of the visualizations used in the study.

## 3. RESULTS AND DISCUSSION

In this section, all of the results were discussed, and appropriate visualizations were also included to backed-up some findings.

### *3.1 Data Exploration*

Due to the nature of the data set, the geometric histogram is produced which indicates that fake news tends to be shorter while real news is longer in nature.

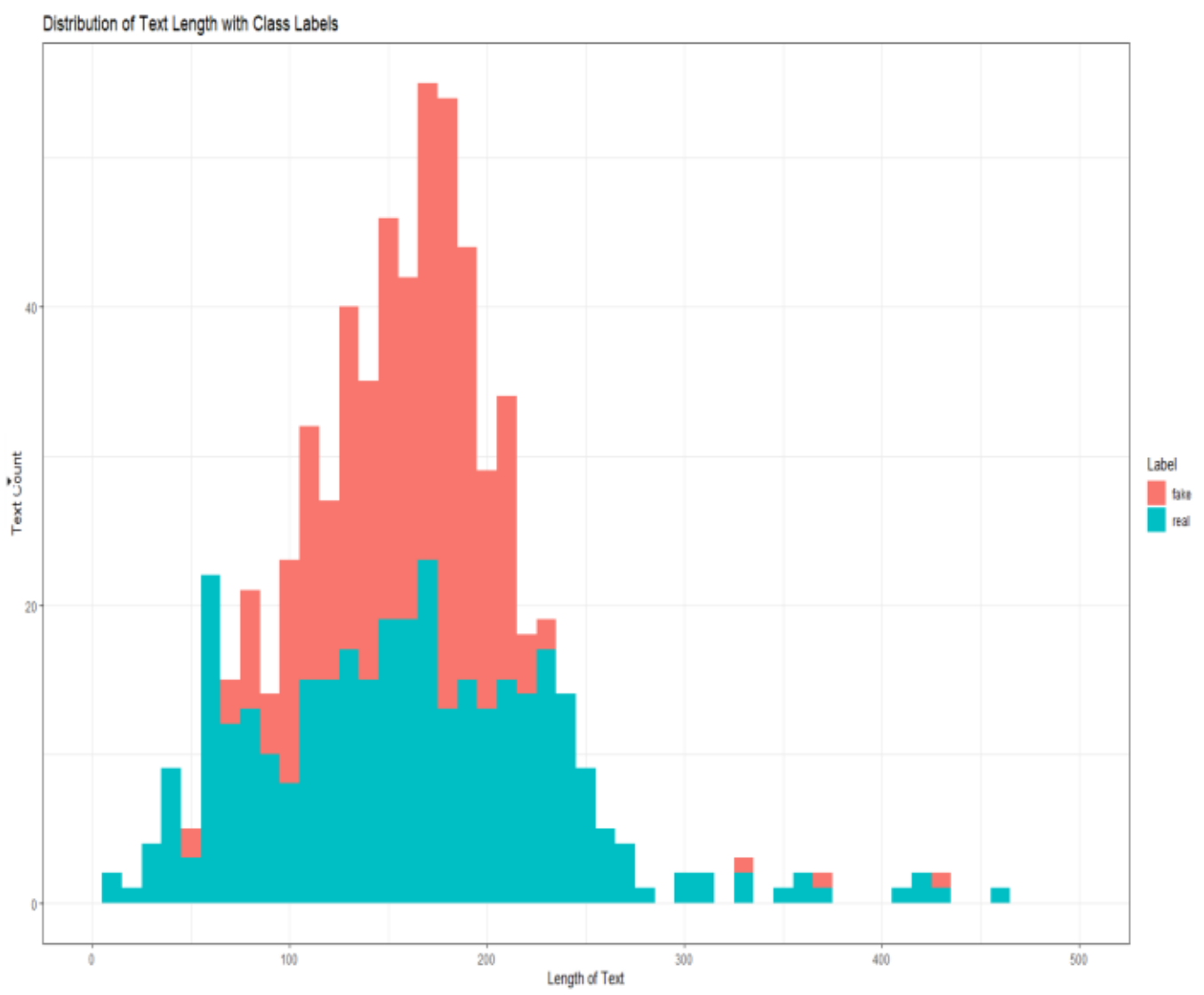

**Figure 3. Geometric histogram of the distribution of text length with class labels**

### *3.2 Term Frequency - Inverse Document Frequency*

Like data exploration, visualization was also made in feature extraction to see the frequent terms after the text pre-processing.

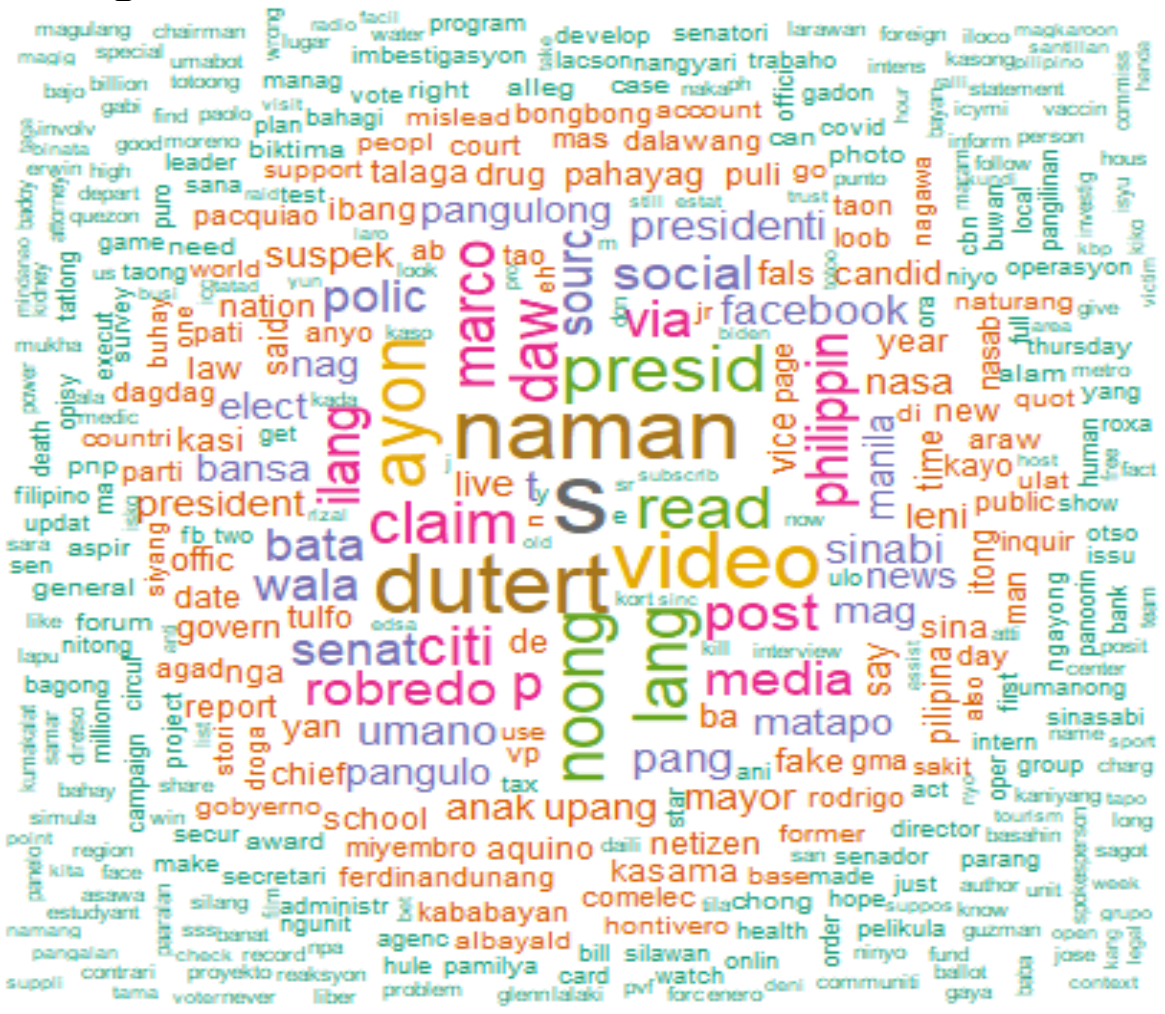

**Figure 4. Word cloud of frequent terms in training data set of the model**

Some of the words in figure 4 is lacking one to two characters in its last part, this is because of the stemming done on the data set. This significantly reduces the features from around 8000 to 7100, which improves furthermore the accuracy of the model. The word cloud used in this visualization accepts all of the tokens to be plotted.

### *3.3 Evaluation*

Same with the Simple Naive Bayes algorithm, Multinomial Naive Bayes is also evaluated using a confusion matrix. The table below shows the confusion matrix for the training of the model as well as the prediction.

**Table 2: Confusion Matrix of the trained model and its predicted values**

| | TP | TN | FP | FN |
|---|---|---|---|---|
| **Training (70%)** | 276 | 275 | 2 | 1 |
| **Test (30%)** | 113 | 97 | 5 | 21 |

Table 1 shows the true positive (TP), true negative (TN), false positives (FP), and false negatives (FN). The trained model shows an accurate prediction due to a low number of FP and FN, while it performs a little bit poorly in predicting unseen values with 5 FP and 21 FN. This directly affects its accuracy which is seen in Table 2, together with precision, recall, and f1 score.

**Table 3: Accuracy, Precision, Recall, and F1 Score of the trained model and its predicted values**

| | Accuracy | Precision | Recall | F1 Score |
|---|---|---|---|---|
| **Training (70%)** | 99.46% | 99.64% | 99.28% | 99.46% |
| **Test (30%)** | 88.98% | 95.76% | 84.33% | 89.68% |

Accuracy is one of the main criteria for evaluation of the model to know its effectivity in successfully predicting values based on features extracted through TF-IDF. Aside from accuracy, precision, recall, and F1 score are also considered. The training dataset has a good evaluation with 99.46%, 99.64%, 99.28%, and 99.46% respectively, while prediction in the test dataset gives a mixed rating of 88.98%, 95.76%, 84.33%, and 89.68%.

## 4. RECOMMENDATIONS AND CONCLUSION

The training data set gives high accuracy of 99.46%, while prediction using the test data set gives an 88.98%. This indicates that the ML model is optimized in training, but has poor generalization in test data, which could lead to overfitting of the model. Overfitting refers to the scenario where a machine learning model can't generalize or fit well on unseen dataset. A clear sign of machine learning overfitting is if its error on the testing or validation dataset is much greater than the error on training dataset [16].

Even if there is a problem with the overfitting of the model, the precision gives a 99.64% and 95.76% which has a high percentage on both datasets. This indicates that the model has a high percentage of correctly identifying a piece of news as fake.

In contrast, the recall indicates that the model has a medium efficiency in labeling and classifying fake news. This is due to the recall of 84.33% of the test data set prediction.

In comparison with accuracy, the F1 Score is the harmony or mean between precision and recall. This is a good evaluation criterion if there is an uneven cost between false negative and false positive. In, here F1 score of 89.68% will be important due to the high number of False negatives in the prediction. Tagging fake news as real news is highly alerting since it could spread and may create a domino impact.

Furthermore, it is recommended that there should be cross-validation done before text pre-processing, to see the improvement before and after the process. Lastly, using other classification algorithms and creating ensemble learning could increase the accuracy in predicting unseen data.